\documentclass[10pt, a4paper]{article}

\usepackage[final]{lrec2026} % this is the new style # review oder final
\usepackage{times}
\usepackage{latexsym}
\usepackage[T1]{fontenc}
\usepackage[utf8]{inputenc}
\usepackage{microtype}
\usepackage{inconsolata}
\usepackage{graphicx}
\usepackage{enumitem}
\usepackage{xcolor}
\usepackage{soul}
\usepackage{multirow}
\usepackage{array}
\usepackage{lscape}
\usepackage{svg}
\usepackage{tcolorbox}
\usepackage{amsmath}
\usepackage{subcaption}
\usepackage{lineno}
\usepackage{longtable}
\usepackage{caption}
\usepackage{subcaption}
\usepackage{xurl}
\usepackage{float}
\usepackage{stfloats}
\usepackage{dingbat}

\usepackage[utf8]{inputenc}
\usepackage[linesnumbered,ruled,vlined]{algorithm2e}
\usepackage{amsmath}

\usepackage{booktabs}        % Für schöne Tabellenlinien (\toprule, \midrule, \bottomrule)
\usepackage{tabularx}        % Für flexiblere Tabellenbreiten (optional)
\usepackage{adjustbox}       % Für Feinjustierungen der Tabellengröße (optional)

\usepackage{siunitx}

\newcommand{\newcheckmark}{\raisebox{0.6ex}{\scalebox{0.7}{$\sqrt{}$}}}
\newcommand{\newcrossmark}{\scalebox{0.85}[1]{$\times$}}

\newcommand{\PreserveBackslash}[1]{\let\temp=\\#1\let\\=\temp}
\newcolumntype{R}[1]{>{\raggedleft\arraybackslash}p{#1}} % Rechtsbündig
\newcolumntype{L}[1]{>{\raggedright\arraybackslash}p{#1}} % Linksbündig
\newcolumntype{C}[1]{>{\centering\arraybackslash}p{#1}} % Zentriert

\usepackage{caption}
\usepackage{colortbl}

\title{AnnoABSA: A Web-Based Annotation Tool for Aspect-Based Sentiment Analysis with Retrieval-Augmented Suggestions}

\name{Nils Constantin Hellwig$^1$, Jakob Fehle$^1$, Udo Kruschwitz$^2$, Christian Wolff$^1$}

\address{
$^1$Media Informatics Group, University of Regensburg, Regensburg, Germany \\
$^2$Information Science Group, University of Regensburg, Regensburg, Germany \\
nils-constantin.hellwig@ur.de, jakob.fehle@ur.de, udo.kruschwitz@ur.de, christian.wolff@ur.de
}

\abstract{We introduce AnnoABSA, the first web-based annotation tool to support the full spectrum of Aspect-Based Sentiment Analysis (ABSA) tasks. The tool is highly customizable, enabling flexible configuration of sentiment elements and task-specific requirements. Alongside manual annotation, AnnoABSA provides optional Large Language Model (LLM)-based retrieval-augmented generation (RAG) suggestions that offer context-aware assistance in a human-in-the-loop approach, keeping the human annotator in control. To improve prediction quality over time, the system retrieves the ten most similar examples that are already annotated and adds them as few-shot examples in the prompt, ensuring that suggestions become increasingly accurate as the annotation process progresses. Released as open-source software under the MIT License, AnnoABSA is freely accessible and easily extendable for research and practical applications.\\ \newline  }

\begin{document}

\maketitleabstract

\section{Introduction}

Aspect-Based Sentiment Analysis (ABSA) constitutes a fine-grained approach to sentiment analysis that goes beyond document-level polarity classification by identifying specific aspects within a text and determining the sentiment orientation associated with each aspect \citep{pontiki2016semeval}. The field encompasses various ABSA subtasks that differ in their granularity of aspect identification. These tasks involve combinations of the following sentiment elements: aspect term \textit{a}, aspect category \textit{c}, opinion term \textit{o}, and sentiment polarity \textit{p}. For instance, in the sentence \textit{``The pizza was delicious.''}, ``pizza'' represents the aspect term, ``food general'' could constitute the associated aspect category, ``delicious'' serves as the opinion term, and the sentiment polarity is positive. In cases where no aspect term is given for an aspect (=implicit aspect), the aspect term is set to ``NULL'', e.g. \textit{``It was delicious.''}. A sentence may contain several aspects, resulting in several aspects that need to be annotated.

Due to the granularity of ABSA, the creation of annotated resources for training and evaluating ABSA-specific models remains highly time-consuming and labour-intensive \citep{nasution2024chatgpt, negi2024hybrid, wang2024generative}. Resources are particularly scarce for low-resource languages and domain-specific contexts \citep{xu2025ds2, le2023problem, fehle2025german}.

The scarcity of annotated datasets is also reflected in the limited availability of specialized annotation tools, with only a few dedicated solutions currently existing. To date, only general-purpose annotation frameworks such as \textit{INCEpTION}\footnote{INCEpTION \citep{klie-etal-2018-inception}: \url{https://inception-project.github.io/}} \citep{wojatzki2017germeval}, \textit{BRAT}\footnote{BRAT: \url{https://github.com/nlplab/brat}} \citep{pontiki2016semeval, pontiki2015semeval, pontiki2014semeval} and \textit{Label Studio}\footnote{Label Studio: \url{https://labelstud.io/}} \citep{hellwig2024gerestaurant, fehle2025german} have been reported to be used for ABSA annotation tasks. However, these lack essential functionalities required for certain subtasks. For example, in Target Aspect Sentiment Detection (TASD), a text may contain numerous implicit aspects that do not correspond to specific textual spans but can be assigned to an aspect category and sentiment polarity. The aforementioned tools cannot handle such dynamic lists of entries, in this case aspect annotations of implicit opinions. 

% Besides general-purpose annotation tools, \citet{hellwig-etal-2025-still} employed Google Sheets\textit{Google Sheets}\footnote{Google Sheets: \url{https://workspace.google.com/products/sheets}} for annotating a dataset for Aspect Sentiment Quad Prediction (ASQP, considers all four sentiment elements aspect term, aspect category, opinion term and sentiment polarity). Each aspect was placed in a separate cell, with the sentiment elements being separated by semicolons. All sentiment elements were inserted in the cells through transcription or copy-paste operations. This approach required manual transcription or copy-paste operations for all sentiment elements, resulting in substantial time overhead and necessitating a subsequent validation procedure to ensure label integrity, specifically, a verification that annotated phrases correspond exactly to their textual occurrences and that only predefined aspect categories were used.

Given the challenges associated with manual data annotation, recent research has increasingly explored Large Language Models (LLMs) to minimize annotation effort across various NLP domains, including social media analysis \citep{hasan2024zero, mu2023navigating, zhang2024sentiment}, (bio-)medicine \citep{labrak2024zero, ateia2023chatgpt}, and finance \citep{deusser2024leveraging, deng2024leveraging}. For ABSA, few-shot learning has demonstrated competitive performance in ABSA tasks, achieving micro-averaged F1 scores approaching those of fine-tuned models while requiring only a few examples \citep{hellwig-etal-2025-still, zhou2024comprehensive}. However, these approaches still fall short of the performance levels reported for models specifically fine-tuned for ABSA tasks \citep{hellwig-etal-2025-still, zhang2024sentiment, zhou2024comprehensive}. 

In this work, we introduce \textbf{AnnoABSA}, a web-based annotation tool designed for ABSA. The tool provides extensive customization capabilities and supports all major ABSA subtasks documented in the literature \citep{zhang2022survey}, including aspect term extraction, aspect category classification, sentiment polarity detection, opinion term identification, and their various combinations (pairs, triplets, quadruples).

Following recent studies in other time-intensive annotation tasks \citep{ghazouali2025visiofirm, kim2024meganno, sahitaj2025acl, li2025softwarex}, we aimed to integrate the capabilities of foundational models to assist annotators in the annotation process. Beyond traditional manual annotation functionality, AnnoABSA optionally incorporates a Retrieval Augmented Generation (RAG)-based suggestion mechanism that combines the strengths of LLM-based predictions with human expertise. Our proposed RAG mechanism retrieves the most semantically similar examples from the pool of instances previously annotated during the annotation process to guide the LLM in providing suggestions. This hybrid approach leverages the efficiency and consistency of large language models while preserving the nuanced judgment and domain knowledge of human annotators, thereby balancing annotation speed with quality.

Our main contributions are as follows:

\begin{itemize}
    \item We present the first open-source annotation tool for ABSA with comprehensive compatibility across all ABSA subtasks and release it under the permissive MIT licence at \url{https://github.com/NilsHellwig/AnnoABSA}.
    \item We introduce a retrieval-augmented LLM-based suggestion mechanism that leverages the most similar annotated examples to enhance annotation efficiency while improving suggestion quality over time.
    \item We demonstrate through systematic evaluation that RAG-based suggestions significantly outperform random sampling baselines in terms of prediction performance.
    \item We provide evidence from a controlled study showing that expert annotators achieve a statistically significant reduction in annotation time (30.51\%) when assisted by RAG-based suggestions compared to unassisted manual annotation.
\end{itemize}

% TODO: Problem Label Studio: Implicit schwer abbilden
% TODO: Anmerken, dass es teils auch nicht dokumentiert wurde, womit annotiert wird also wenig aufmerksamkeit!
% TODO: bei Contributions lightweight model erwähnen
% TODO: Performance konkret nennen (?)
% TODO: Tabelle erstellen.
% TODO: Beschreiben, dass viele annotationen nötig, hier evlt auf bild tool referenzieren

% ----
% TODO zeitgain reporten von bildannotation tool?
% TODO: BRAT nur python 2
% TODO: Relations nicht zwingen, fehler denkbar
% TODO: Ganze Tokens prüfen
% TODO: Tool aktualität python version, letztes update etc.
% TODO: LLM Format: Model oder LLM?
% TODO: LLM-Predictions in gewünschtem format
\section{System Description}

\begin{table*}[ht]
\centering
\resizebox{0.95\textwidth}{!}{%
\begin{tabular}{l|c|c|c|c}
\toprule
\textbf{Feature} & \textbf{Label Studio} & \textbf{INCEpTION} & \textbf{BRAT} & \textbf{AnnoABSA} \\
\midrule
\multicolumn{5}{l}{\textit{Technical requirements}} \\
\midrule
\rowcolor{gray!5}
Open source & \newcheckmark\ (Apache 2.0) & \newcheckmark\ (Apache 2.0) & \newcheckmark\ (MIT) & \newcheckmark\ (MIT) \\
Installation & Docker, Python 3 & Docker, Tomcat, Java & Python 2 & Python 3 \\
\rowcolor{gray!5}
Regular updates & \newcheckmark & \newcheckmark & \newcrossmark\ Oct. 2021 & \newcheckmark \\
Web-based & \newcheckmark & \newcheckmark & \newcheckmark & \newcheckmark \\
\rowcolor{gray!5}
Multi-user support & \newcheckmark & \newcheckmark & \newcrossmark & \newcheckmark\\
\midrule
\multicolumn{5}{l}{\textit{User interface \& usability}} \\
\midrule
Annotation setup & XML template & Project setup & Project setup & JSON config/CLI \\
\rowcolor{gray!5}
Interface design & Highly customizable & Technical, functional & Technical, functional & Modern, intuitive \\
\rowcolor{gray!5}& (XML template) & & & \\
Language translation & \newcheckmark & \newcrossmark & \newcrossmark & \newcheckmark \\
\rowcolor{gray!5}
Annotation guidelines & \newcheckmark\ Configurable & \newcrossmark & \newcrossmark & \newcheckmark\ Integrated in the \\
\rowcolor{gray!5} & with XML & & & interface as PDF \\
Documentation & \newcheckmark & \newcheckmark & \newcheckmark & \newcheckmark \\
\rowcolor{gray!5} Support & Community forum, & GitHub issues & GitHub issues & GitHub issues, \\
\rowcolor{gray!5} & GitHub issues & & (inactive developers) & email \\
\midrule
\multicolumn{5}{l}{\textit{Data management}} \\
\midrule
\rowcolor{gray!5}
Import/Export formats & JSON, TXT, XML; & UIMA, TSV, JSON, TXT & TXT, ANN & JSON, CSV \\
\rowcolor{gray!5} & CSV export only & & (BRAT stand-off) & \\
\midrule
\multicolumn{5}{l}{\textit{General annotation functionalities}} \\
\midrule
\rowcolor{gray!5}
Multi-labelling functionality & \newcheckmark & \newcheckmark & \newcheckmark & \newcheckmark \\
Relationship tagging & \newcheckmark & \newcheckmark & \newcheckmark & \newcheckmark \\
\rowcolor{gray!5} Label customization & \newcheckmark\ XML & \newcheckmark\ Project & \newcheckmark\ Config & \newcheckmark\ JSON \\
\rowcolor{gray!5} & template & configuration & files & configuration \\
 Team collaboration & \newcheckmark\ Full support & \newcrossmark & \newcrossmark\  & \newcrossmark \\
 & with role management &  &  &  \\
\rowcolor{gray!5} Token identification\textsuperscript{1} & \newcheckmark & \newcheckmark & \newcheckmark & \newcheckmark \\
\midrule
\multicolumn{5}{l}{\textit{ABSA-specific features}} \\
\midrule
\rowcolor{gray!5}
Validation\textsuperscript{2} & \newcrossmark & \newcrossmark & \newcrossmark & \newcheckmark \\
AI-based suggestions & \newcheckmark\ Integration for & \newcheckmark\ Integration for & \newcrossmark & \newcheckmark \\
& individual models/APIs & individual models/APIs & & \\
\rowcolor{gray!5}
Dynamic lists\textsuperscript{3} & \newcrossmark & \newcrossmark & \newcrossmark & \newcheckmark \\
Assigning (multiple)  &  &  &  &  \\
categories to a & \newcrossmark & \newcrossmark & \newcrossmark & \newcheckmark \\
text span  &  &  &  &  \\
\bottomrule
\end{tabular}
}
\caption{\textbf{Comparison of Annotation Tools for ABSA.} Detailed comparison of four annotation platforms across technical requirements, usability features, data management capabilities, and ABSA-specific functionalities. The table constitutes an extension of the comparison provided by \citet[p.~359]{colucci2024text}, who evaluated non-semantic textual annotation tools based on the following criteria: multi-labelling functionality, annotation suggestions, relationship tagging, label customization, and team collaboration.}
\label{tab:annotation-tools}
\footnotesize
\begin{flushleft}

\textsuperscript{1} \textit{Token Identification}: Automatic identification of token boundaries in text to prevent annotation errors where spans begin or end within words.

\textsuperscript{2} \textit{Validation}: Label Studio, INCEpTION, and BRAT do not enforce mandatory linking of sentiment elements within aspect tuples. For example, in the TASD task (tuples consisting of aspect term, aspect category, and sentiment polarity), annotators may mark an aspect term but forget to set the corresponding aspect category and sentiment polarity. AnnoABSA prevents such errors through strict validation, ensuring each tuple contains exactly the required number of sentiment elements for a specific task. Individual sentiment elements are also validated (e.g., sentiment polarity must be "positive", "negative", or "neutral"; aspect categories must be from predefined lists; aspect/opinion terms must be either NULL for implicit aspects/opinions or substrings of the text that should be annotated).

\textsuperscript{3} \textit{Dynamic Lists}: Annotations for ABSA tasks may comprise an unlimited number of aspects. As an example, for the ACD task (which considers pairs of aspect category and sentiment polarity), a tool should offer the functionality to assign unlimited combinations of aspect categories and sentiment polarities to documents. Label Studio, INCEpTION, and BRAT cannot handle unlimited assignments of categorical variable combinations to documents.

\end{flushleft}

\end{table*}

\begin{figure*}[!h]
    \centering
    \fbox{\includegraphics[width=0.97\textwidth]{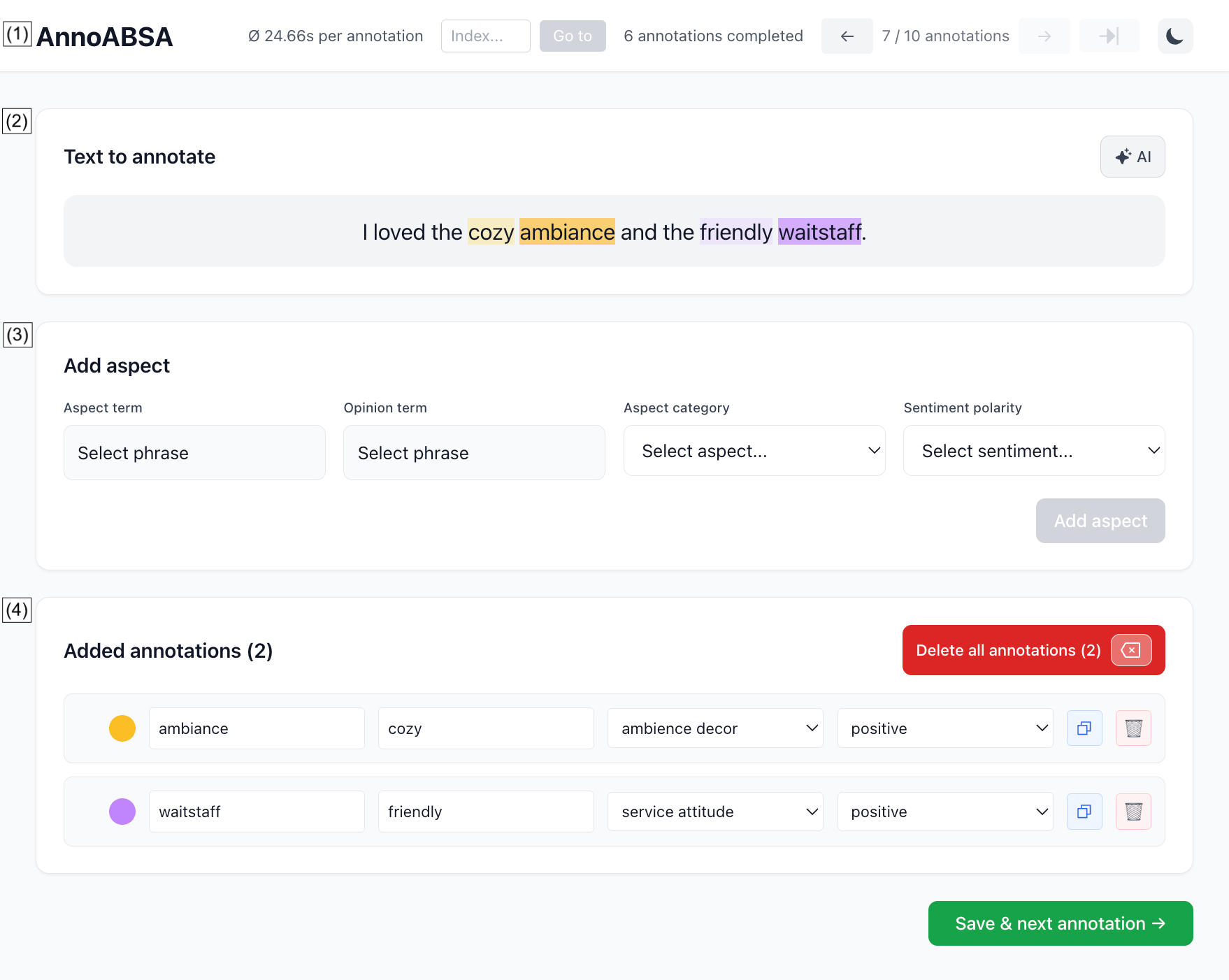}}
    \caption{\textbf{UI of AnnoABSA.} The UI consists of four components: (1) top navigation bar to change the currently displayed example, (2) annotated text with highlighted sentiment annotations, (3) aspect addition panel, and (4) editable list of added annotations.}
    \label{fig:ui}
\end{figure*}

\subsection{Motivation}

We present AnnoABSA, a novel annotation tool designed to address the need to support all ABSA tasks (see Appendix~\ref{appendix:absa_tasks}) while providing an accessible and intuitive user interface (UI) that facilitates efficient annotation with minimal user interaction. In this section, we detail the system architecture, UI design decisions, and AnnoABSA's data management. Additionally, Table \ref{tab:annotation-tools} characterizes the core features of AnnoABSA and compares them with existing annotation tools previously utilized for ABSA tasks.

% easy code adaption

\subsection{System Architecture}

AnnoABSA is implemented using \textit{React.js}\footnote{React.js: \url{https://reactjs.org/}}, a frontend framework specifically designed for Single-Page Applications (SPAs). We utilized \textit{TypeScript}\footnote{TypeScript: \url{https://www.typescriptlang.org/}} as the primary programming language for the frontend, a statically-typed superset of JavaScript that enhances code robustness and reduces error susceptibility. The backend employs \textit{FastAPI}\footnote{FastAPI: \url{https://fastapi.tiangolo.com/}}, a Python RESTful API framework that enables the integration of essential Python packages such as \textit{Pandas}\footnote{Pandas: \url{https://pandas.pydata.org/}} and \textit{NumPy}\footnote{NumPy: \url{https://numpy.org/}} for efficient data processing and manipulation.

AnnoABSA can be started through a command-line interface by providing the text to be annotated in either JSON or CSV format, along with a configuration file specifying the annotation parameters.

\begin{tcolorbox}[colback=gray!10, colframe=gray!50, arc=2mm]
\texttt{\$ ./annoabsa reviews.json --load-config config.json}
\end{tcolorbox}

\vspace{0.2cm}

The system offers extensive customization options that can be specified in a configuration file or through CLI flags. These configuration parameters include the input text file for annotation, the sentiment elements to be considered (aspect terms, aspect categories, opinion terms, sentiment polarity), a list of valid sentiment polarities and aspect categories, boolean options for specifying if implicit aspect terms and/or opinion terms are valid. Hence, AnnoABSA supports the annotation of text data in any language. A comprehensive overview of all supported flags is provided in Appendix~\ref{appendix:flags}. After executing the CLI tool, the AnnoABSA UI is opened in the web browser.

% Wie wird anwendung gestartet grob (CLI)
% TODO: CLI flags einfügen tabelle
% TODO: Links für react etc einfügen

\subsection{UI Design}

The UI is presented in Figure \ref{fig:ui}. The interface components were designed for comfortable use on both desktop computers and tablets. Following minimalist design principles \citep{meyer2015minimalism}, we focused on the main content, avoided redundant visual elements, and employed flat design aesthetics.

\subsubsection{Topbar}

The topbar provides a convenient navigation between examples, using arrow keys to move to the previous or next instance. Additionally, a double-arrow button on the right allows users to jump directly to the unannotated example with the highest index, and an input field enables navigation to a specific index. % Optional features (configurable via CLI flags) include a dark mode toggle (moon icon) similar to LabelStudio and a display of the average annotation speed.

\subsubsection{Form to Add New Aspect}

The subsequent section in the UI enables adding new aspect annotations. Depending on the CLI specification, varying numbers of sentiment elements to be annotated are displayed. Figure \ref{fig:ui} shows all four elements, though configurations with fewer sentiment elements can also be configured. Categorical sentiment elements, aspect category and sentiment polarity, can be selected using dropdown menus, while aspect term and opinion term each trigger a popup interface.

\begin{figure}[h]
    \centering
    \includegraphics[width=0.48\textwidth]{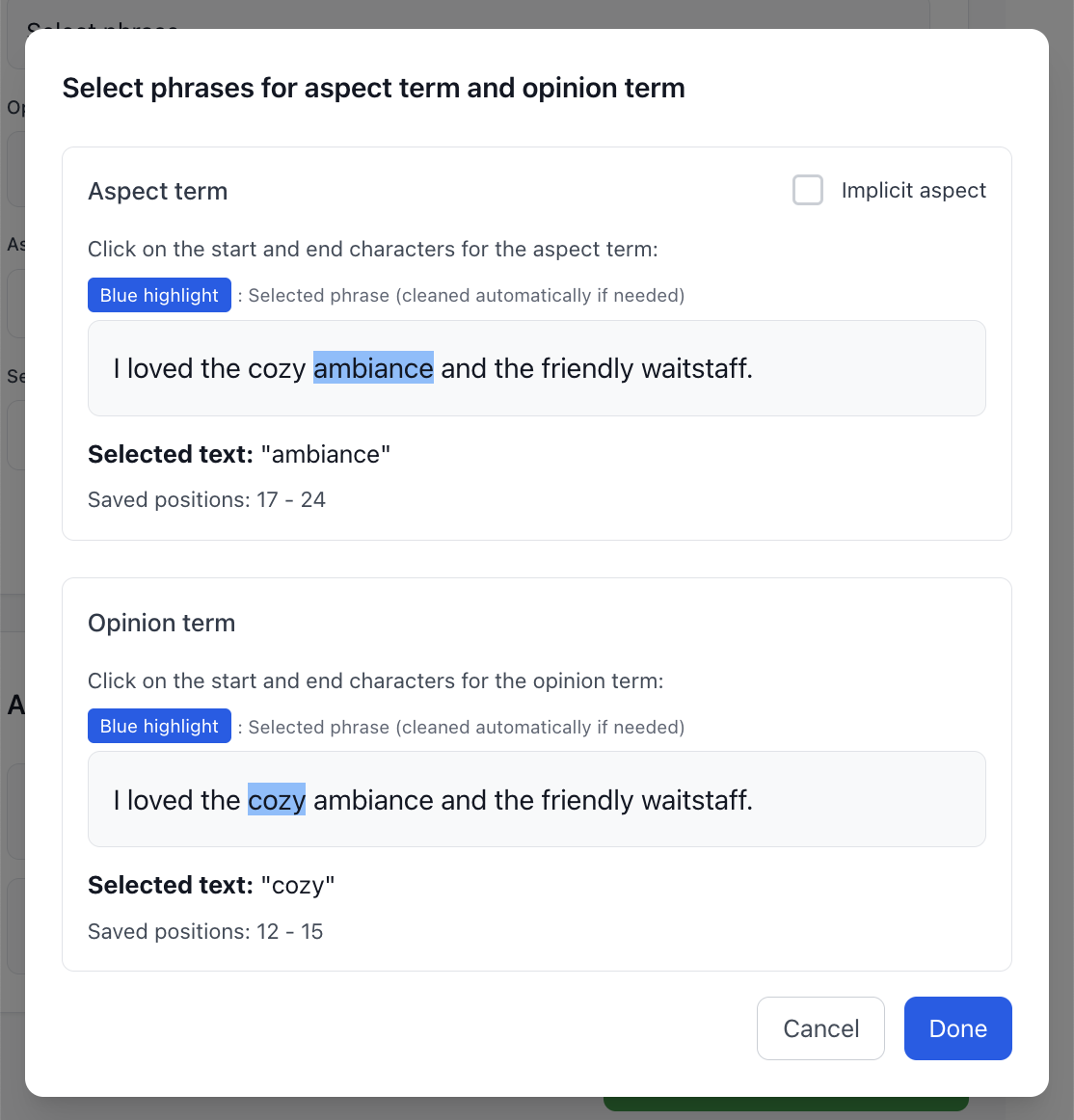}
    \caption{\textbf{Popup to add or manipulate phrase annotations.} In case both aspect term and opinion term annotations are required, the text is displayed twice, once for each phrase annotation. Implicit aspects can be marked using a checkbox.}
    \label{fig:ui-popup}
\end{figure}

As shown in Figure \ref{fig:ui-popup}, the popup displays the text to be annotated twice, once for aspect term annotation and once for opinion term annotation. When only one phrase type needs to be annotated, the text is displayed once. We considered implementing a tool-switching approach (separate selectable tools for aspect term and opinion term selection), but this would have required additional clicks. Our chosen interface requires only phrase marking and clicking ``Done'', thus minimizing user actions to only those absolutely necessary.

Phrase boundaries are defined by clicking on individual characters or, if configured, tokens. Once a phrase is marked, both the text and position are displayed for confirmation. Notably, this popup appears both when annotating new aspects and when editing existing ones. After all sentiment elements are specified, users click ``Add aspect'' to append the sentiment element to the list of applied annotations presented below the ``Add aspect'' section.

\subsubsection{Annotation List}

All annotated aspects are displayed in a list and can be modified at any time. We included a duplication button for aspects, designed to assist in cases where aspects differ in only one sentiment element. For example, in the sentence \textit{``The pizza and the burger were delicious''}, the Aspect Sentiment Quad Prediction (ASQP) gold label consists of two quadruples: \texttt{('pizza', 'food quality', 'delicious', 'positive')} and \texttt{('burger', 'food quality', 'delicious', 'positive')}. During annotation, one quadruple can be duplicated and the relevant sentiment element modified accordingly. Annotations can be deleted as needed.

%TODO: Wenige Klicks erforderlich

\subsection{Data Handling}

The CSV or JSON file containing all examples to be annotated is directly modified during the annotation process, with annotations being appended as the annotation work progresses. In addition to the sentiment elements and phrase positions within the given text, annotation duration can optionally be stored. The use of JSON format enables straightforward integration for NoSQL databases if required, with minimal code modifications.

% wieso easily extendable
% TODO: Guidelines einfügen
\section{Retrieval Augmented Suggestions}

Given the absence of annotated training examples for supervised models in a situation where a human annotator starts annotating text for ABSA, we adopt an LLM-based approach for generating annotation suggestions. Previous research \citep{hellwig-etal-2025-still, zhou2024comprehensive, zhang2024sentiment} demonstrated that LLMs utilizing a small, fixed set of few-shot examples achieve performance approaching fine-tuned models, particularly in low-resource scenarios, and, that Retrieval Augmented Generation (RAG)-based few-shot learning achieved higher performance scores than random sampling across various ABSA tasks.

This section examines technical considerations including model support and prompting techniques, followed by a comparative evaluation of random and RAG-based sampling approaches for annotation suggestion generation.

\subsection{Model Support}

AnnoABSA provides flexible model integration through appropriate CLI parameters, supporting both commercial models via the OpenAI API\footnote{OpenAI API: \url{https://openai.com/api/}} and open-source models through the locally hosted Ollama API\footnote{Ollama: \url{https://ollama.com/}}. These toolkits were selected based on their platform independence and native implementation of structured output capabilities. Structured outputs via guided decoding ensure that LLM suggestions are constrained to the aspect categories and sentiment polarities defined in the CLI configuration, preventing the generation of hallucinated labels. Furthermore, structured outputs guarantee that predicted phrases are present in the text that needs to be annotated.

\subsection{Prompt}

We adopted the prompt structure employed by \citet{hellwig-etal-2025-still} and \citet{gou2023mvp}, which comprises a task description, demonstrations, and the text to be annotated. A modification of the prompt involves formatting the demonstration's labels in JSON, similar to the aforementioned enforced structured output. An example of the prompt template is provided in Appendix~\ref{appendix:prompt}.

\subsection{Demonstration Selection Strategy Evaluation}

\begin{figure*}[ht]
    \centering
    \includegraphics[width=\textwidth]{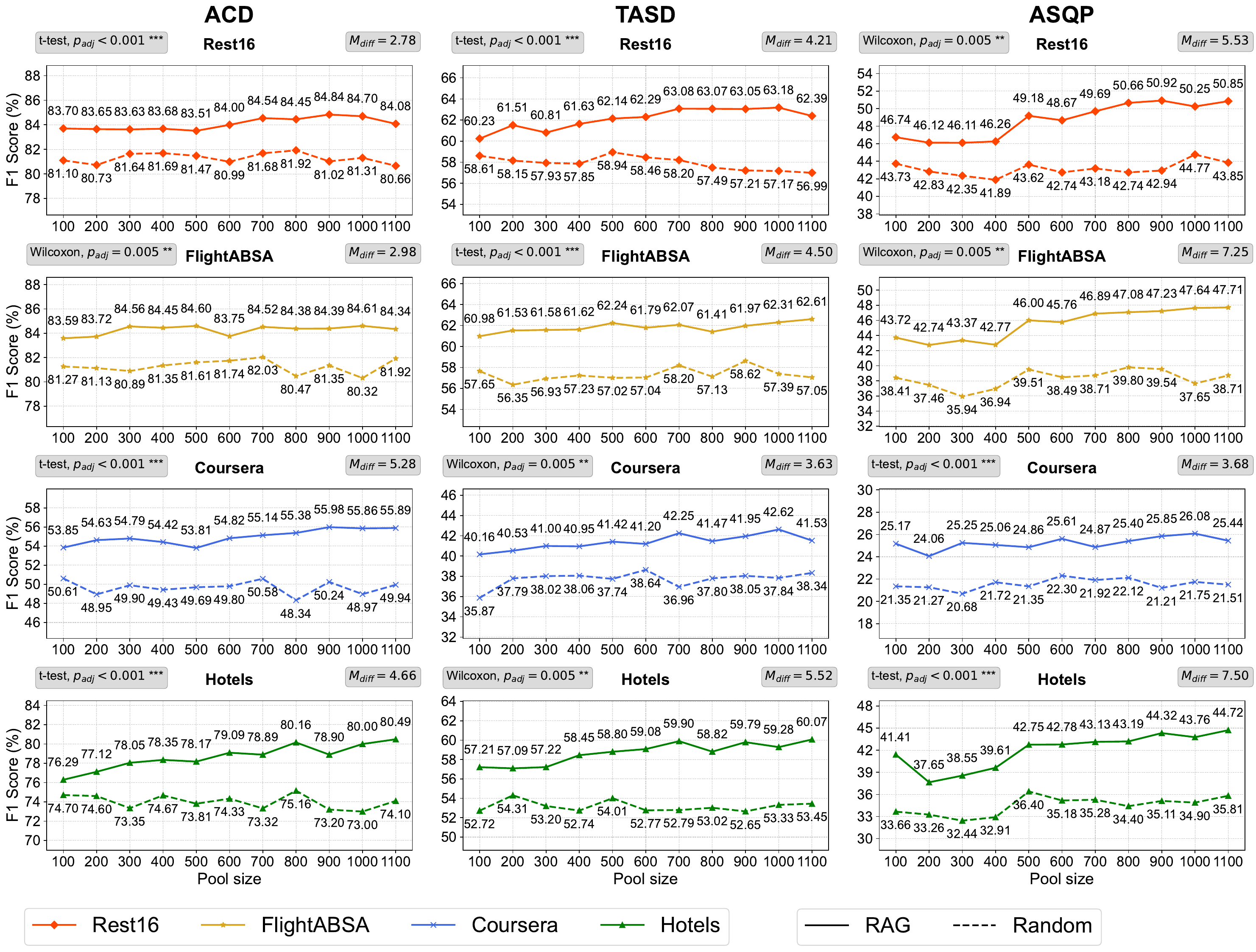}
    \caption{\textbf{F1 score comparison of RAG-based and random sampling approaches.} RAG consistently outperforms random sampling across all configurations. Statistical tests (paired t-test or Wilcoxon signed-rank test with Holm-Bonferroni correction) confirm all differences are significant. $M_{diff}$ shows mean performance differences.}
    \label{fig:rag-performance}
\end{figure*}

Although the RAG-based approach by \citet{zhou2024comprehensive} which utilized the \textit{k} most semantically similar training examples as few-shot demonstrations achieved higher performance scores than random selection, their methodology cannot be directly transferred to the annotation process. Their approach supposes access to a fully annotated training corpus for example selection, which contrasts with annotation scenarios where the set of human-annotated instances (pool) expands during the labelling process. To address this discrepancy, we conducted a performance analysis between random sampling and RAG-based sampling strategies within an annotation framework. All LLM executions in our performance analysis were conducted on an NVIDIA RTX PRO 6000 GPU with 96 GB VRAM.

\subsubsection{Methodology}

\textbf{Selection Strategy.} Following the methodology established by \citet{zhou2024comprehensive}, we employ BM25 \citep{robertson2009probabilistic} as a sparse retrieval algorithm for RAG, as it enables rapid similarity comparisons, thereby facilitating fast suggestion generation. For random sampling, the examples are randomly selected from the pool.

\textbf{Pool Size.} To investigate performance evolution as the number of gold-labelled examples available for few-shot demonstration retrieval increases, we analysed a spectrum ranging from 0 to 1,100 available training examples in incremental steps of 100. For instance, when the pool size is 300, the few-shot examples are selected from 300 examples. We considered 1,100 examples as the maximum pool size, as this was the largest multiple of 100 available across all datasets.

\textbf{Datasets.} Performance was evaluated across four datasets spanning diverse review domains. These datasets encompass reviews on restaurants (SemEval 2016, Rest16) \citep{pontiki2016semeval, zhang2021aspect}, e-learning courses (Coursera) \citep{chebolu2024oats}, airlines (FlightABSA) \citep{hellwig-etal-2025-still}, and hotels \citep{chebolu2024oats}. We randomly selected 1,100 examples from the respective training sets.

\textbf{LLM Configuration.} Similar to \citet{hellwig-etal-2025-still}, we employed \textit{Google}'s Gemma-3-27B \citep{team2025gemma} with the temperature set to 0, ensuring deterministic selection of the highest probability token during next-token prediction. The prompt context incorporated 10 few-shot examples. For each combination of pool size, task and dataset, the LLM was executed five times. Each time, a different random seed was employed to ensure varied selections of the 1,100 examples extracted from the respective training set.

\textbf{Tasks.} We evaluated three tasks of varying complexity: one single-aspect task (Aspect Category Detection, ACD) and two tuple prediction tasks: Target Aspect Sentiment Detection (TASD), and Aspect Sentiment Quad Prediction (ASQP). ACD requires the identification of all aspect categories addressed within a given text. TASD extracts opinion triplets comprising aspect term, aspect category, and sentiment polarity, while ASQP additionally extracts opinion terms, representing the most fine-grained ABSA task.

\textbf{Evaluation metrics.} Evaluation for each step was performed on the full respective test set of each dataset. As common in the field of ABSA, the reported evaluation metric is the micro-averaged F1 score \citep{zhang2022survey}. We publish all predicted labels and provide performance scores of the macro-averaged F1 score, precision, and recall in our GitHub repository.

\subsubsection{Results \& Discussion}

The results (see Figure \ref{fig:rag-performance}) demonstrated that the RAG approach consistently outperformed random few-shot selection across all pool sizes and tasks for all datasets. Notably, RAG achieved substantial performance gains, with differences ($M_{diff}$) of up to six percentage points compared to random sampling in several instances.

\begin{table}[ht]
\centering
\vspace{0.5em}
\resizebox{0.48\textwidth}{!}{%
\renewcommand{\arraystretch}{1.3}
\Large
\begin{tabular}{
r|*{6}{>{\centering\arraybackslash}p{1.8cm}}
}
\toprule
\multicolumn{1}{c|}{\cellcolor{gray!10}\textbf{\Large Pool}} & 
\multicolumn{6}{c}{\cellcolor{gray!10}\textbf{\Large LLM Execution time (seconds)}} \\
\cmidrule(lr){1-1} \cmidrule(lr){2-7}
& \multicolumn{2}{c|}{\textbf{\Large ACD}} & \multicolumn{2}{c|}{\textbf{\Large TASD}} & \multicolumn{2}{c}{\textbf{\Large ASQP}} \\
\cmidrule(lr){2-3} \cmidrule(lr){4-5} \cmidrule(lr){6-7}
& \textbf{RAG} & \textbf{Random} & \textbf{RAG} & \textbf{Random} & \textbf{RAG} & \textbf{Random} \\
\midrule
\rowcolor{gray!5}
\textbf{100} & 0.816 & 0.851 & 1.292 & 1.240 & 1.708 & 1.622 \\
\textbf{200} & 0.848 & 0.852 & 1.291 & 1.204 & 1.723 & 1.626 \\
\rowcolor{gray!5}
\textbf{300} & 0.846 & 0.871 & 1.291 & 1.211 & 1.736 & 1.642 \\
\textbf{400} & 0.862 & 0.856 & 1.298 & 1.221 & 1.743 & 1.622 \\
\rowcolor{gray!5}
\textbf{500} & 0.858 & 0.867 & 1.314 & 1.220 & 1.743 & 1.647 \\
\textbf{600} & 0.868 & 0.852 & 1.308 & 1.219 & 1.745 & 1.630 \\
\rowcolor{gray!5}
\textbf{700} & 0.864 & 0.884 & 1.311 & 1.210 & 1.739 & 1.640 \\
\textbf{800} & 0.865 & 0.872 & 1.320 & 1.210 & 1.755 & 1.638 \\
\rowcolor{gray!5}
\textbf{900} & 0.868 & 0.864 & 1.322 & 1.204 & 1.759 & 1.635 \\
\textbf{1,000} & 0.872 & 0.856 & 1.310 & 1.202 & 1.745 & 1.636 \\
\rowcolor{gray!5}
\textbf{1,100} & 0.866 & 0.859 & 1.296 & 1.199 & 1.756 & 1.617 \\
\midrule
\textbf{AVG} & 0.858 & 0.862 & 1.305 & 1.213 & 1.741 & 1.632 \\
\bottomrule
\end{tabular}%
}
\vspace{0.3em}
\caption{\textbf{LLM inference time comparison for ABSA tasks.} Average execution time per prediction (in seconds) comparing RAG-based versus random sampling strategies across varying pool sizes for ACD, TASD, and ASQP tasks.}
\label{tab:llm_output_time}
\end{table}

Statistical significance was tested using paired t-tests for normally distributed differences and Wilcoxon signed-rank tests for non-normally distributed sets, followed by Holm-Bonferroni correction \citep{holm1979simple} for multiple testing across 12 comparisons ($\alpha = 0.05$). For each task-dataset combination, we compared the 11 performance scores (one at each pool size) obtained under RAG versus random sampling. Statistical significance was observed across all task-dataset combinations.

Overall, our findings demonstrated that a RAG-based approach achieved higher performance scores than random sampling in the context of a growing pool from which few-shot examples are drawn, with statistically significant differences observed across all tasks and datasets. Accordingly, the RAG approach was integrated into the final version of AnnoABSA. 

Further performance improvements could potentially be achieved by incorporating a larger number of few-shot examples, as previously demonstrated for random sampling by \citet{hellwig-etal-2025-still}. However, such improvements would come at the cost of increased prediction latency, as additional tokens would need to be loaded into the model context. On our NVIDIA RTX PRO 6000 hardware, predictions for individual tasks were completed within seconds, as shown in Table \ref{tab:llm_output_time}, but increasing the number of few-shot examples would proportionally extend processing time or financial costs in the case of a commercial LLM.

\subsection{User Study on Annotation Speed}

As with other LLM-assisted annotation tools, the main motivation behind AnnoABSA's LLM-based suggestions is to reduce annotation time \citep{kim2024meganno}. We therefore conducted a user study to evaluate whether AnnoABSA enables faster annotation compared to manual annotation without AI assistance. We selected ASQP as the target task, which considers the most sentiment elements per tuple among all ABSA tasks (aspect term, aspect category, opinion term, and sentiment polarity) and is therefore the most comprehensive annotation task.

\subsubsection{Methodology}
We conducted a within-subjects user study with 8 expert annotators, comprising four PhD students and four master's students in computer science, all with prior experience in ABSA annotation tasks. The study was conducted in a controlled usability laboratory environment to minimize external distractions, with a research supervisor available exclusively during the initial briefing phase to answer questions before the annotation tasks commenced.  

The study employed a counterbalanced design, where each participant completed two annotation sessions on separate days to eliminate fatigue effects: one session with AI suggestions enabled and one without suggestions. To control for potential ordering and learning effects, we implemented Latin square counterbalancing, systematically varying both the system order (AI-first vs. baseline-first) and dataset assignment (subset A vs. subset B) across participants.

Subsets A and B each consisted of 50 randomly sampled examples from the restaurant dataset published as part of SemEval 2016 Task 5 \citep{pontiki2016semeval}. Participants were provided with the corresponding annotation guidelines for reference. Prior to annotating the 50 test examples in each session, participants completed a familiarization phase with 5 demonstration examples to become acquainted with the tool interface, during which they could ask questions and receive guidance from the research supervisor.

We employed the same LLM (Gemma-3-27B) and GPU configuration as used in our selection strategy evaluation. Given that the time annotators took per example served as our primary evaluation metric, we implemented frontend-based time tracking. The timer was started upon loading each example and terminated when participants opened the subsequent example in the interface.

\subsubsection{Results \& Discussion}

A paired $t$-test revealed a statistically significant difference in mean annotation time per example between the AI-assisted condition ($M = 24.19$, $SD = 3.49$) and the baseline condition ($M = 34.80$, $SD = 6.92$), $t(7) = -3.640$, $p = 0.008$. The Shapiro-Wilk test confirmed that the distribution of differences satisfied the normality assumption ($W = 0.944$, $p = 0.647$). These findings demonstrate that LLM-generated suggestions can substantially accelerate human annotation workflows, yielding a 30.51\% reduction in annotation time.

Our results align with prior work demonstrating that AI-assisted annotation can reduce the temporal demands of human labelling tasks \citep{kim2024meganno, ni2024afacta, sahitaj2025acl, li2025softwarex}, though the reduction observed in our study is more modest than that reported for other tasks. Notably, \citet{ni2024afacta} demonstrated that nearly half of factual claim annotations could be fully automated through consistency-checked GPT-4 outputs, thereby reducing expert annotation effort by approximately 50\%. Similarly, \citet{sahitaj2025acl} reported a 73\% reduction in annotation time for propaganda detection in tweets, while \citet{li2025softwarex} observed a 36\% reduction for AI-assisted pre-annotation of x-ray images.
\section{Conclusion \& Future Work}

We presented AnnoABSA, the first open-source, web-based annotation tool supporting all subtasks of ABSA. AnnoABSA features an intuitive and customizable interface with built-in validation mechanisms and optional LLM-powered suggestions via few-shot prompting. Our evaluation demonstrated that the integrated RAG-based approach, which dynamically leverages the growing pool of annotated examples during the annotation process, significantly outperformed random sampling. A user study revealed significant reductions in annotation time when AI-assisted suggestions are employed.

While this work demonstrates AnnoABSA's functional capabilities and its positive impact on annotation efficiency, future research should systematically investigate how AI assistance affects annotation time, quality, and annotator confidence in ABSA tasks. Such investigations could consider both crowd-sourced workers and domain experts with established ABSA expertise.

Finally, we note that our approach requires no task-specific model training and provides immediate utility upon deployment, making it readily adaptable to diverse NLP annotation scenarios beyond ABSA. 

We invite the community to contribute to AnnoABSA's continued development through our GitHub repository. Feature requests in the form of issues and pull requests are welcome, with our commitment to timely integration of valuable contributions to benefit the broader research community.

% Zusammenfassung was AnnoABSA ist
% weitere änderungen über Issues / Pull Requests and Future Support support
% Was macht RAG Ansatz
% RAG ist bias -> Want to highlight that it's optional
% RAG detailierter Evaluieren. Nutzerstudie sehr klein, größere menge an beispielen, nur für einen ersten einblick
% 

% FW: Active Learning
% Futurue work: adaption für crowd working, limit: wir haben perfekte performance analysier
% Future Work: Active learning
% Future Work: Predictions noch schneller, vlt nicht nur zur Laufzeit, andere LLMs, abhängig von LLM, 

% FW: Kleine LLMs
\section{Ethics Statement and Limitations}

This research was conducted without industrial funding or commercial sponsorship. We employed AI coding agents Claude Sonnet 4.5\footnote{Claude Sonnet: \url{https://www.anthropic.com/claude/sonnet}} and OpenAI's open-source LLM gpt-oss:20b\footnote{gpt-oss:20b: \url{https://ollama.com/library/gpt-oss:20b}} for programming support. Claude Sonnet 4.5 was also used to assist in the formulation of this publication.

Several limitations should be considered when interpreting our results. First, our evaluation of LLM-based suggestions was restricted to Gemma-3-27B due to computational constraints. Larger language models or increased few-shot example sizes could potentially yield superior suggestion quality, albeit at higher computational or financial costs in the case of commercial proprietary LLMs. We executed 580,690 prompts for the evaluation of LLM-based suggestions, which would have incurred substantial costs when employing commercial models. However, AnnoABSA supports arbitrary Ollama and OpenAI-compatible models if one wishes to employ those.

Finally, while we demonstrated significant reductions in annotation time, our evaluation was limited to a small-scale annotation task. The generalizability of these efficiency gains to large-scale real-world scenarios involving thousands of examples remains to be validated, particularly regarding the potential compounding effects of annotator fatigue over extended annotation sessions.

\nocite{*}
\section{Bibliographical References}\label{sec:reference}

\bibliographystyle{lrec2026-natbib}
\bibliography{lrec2026-example}

\onecolumn
\appendix

\section{ABSA Tasks}\label{appendix:absa_tasks}

\begin{table*}[!h]
\centering
\vspace{0.5em}
\resizebox{1.0\columnwidth}{!}{
\renewcommand{\arraystretch}{1.2}
\Large
\begin{tabular}{llll}
\toprule
\textbf{Task Type} & \textbf{Task Name} & \textbf{Elements} & \textbf{Gold Label (Example)} \\
\midrule
\rowcolor{gray!5}
\multirow{3}{2cm}{\textbf{Single}} 
& Aspect Term Extraction (ATE) & \textit{a} & \texttt{['waiter']} \\
& Aspect Category Detection (ACD) & \textit{c} & \texttt{['food quality', 'service speed']} \\
\rowcolor{gray!5}
& Opinion Term Extraction (OTE) & \textit{o} & \texttt{['delicious', 'way too slow']} \\
\midrule
\multirow{10}{2cm}{\textbf{Compound}} 
& Aspect Sentiment Classification (ASC) & \textit{a}, \textit{s} & \texttt{[('waiter', 'negative')]} \\
\rowcolor{gray!5}
& Aspect-Opinion Pair Extraction (AOPE) & \textit{a}, \textit{o} & \texttt{[('waiter', 'way too slow')]} \\
& End-to-End ABSA (E2E-ABSA) & \textit{a}, \textit{s} & \texttt{[('NULL', 'positive'), ('waiter', 'negative')]} \\
\rowcolor{gray!5}
& Aspect Category Sentiment Analysis (ACSA) & \textit{c}, \textit{s} & \texttt{[('food quality', 'positive'), ('service speed', 'negative')]} \\
%\vspace{0.2cm}
%\arrayrulecolor{gray}\cline{2-4} \arrayrulecolor{black}
% \vspace{0.2cm}
& Aspect Sentiment Triplet Extraction (ASTE) & \textit{a}, \textit{o}, \textit{s} & \begin{tabular}[c]{@{}l@{}}\texttt{[('NULL', 'delicious', 'positive'),} \\ \texttt{\ ('waiter', 'way too slow', 'negative')]}\end{tabular} \\
\rowcolor{gray!5}
& Target Aspect Sentiment Detection (TASD) & \textit{a}, \textit{c}, \textit{s} & \begin{tabular}[c]{@{}l@{}}\texttt{[('NULL', 'food quality', 'positive'),} \\ \texttt{\   ('waiter', 'service speed', 'negative')]}\end{tabular} \\
& Aspect Sentiment Quad Prediction (ASQP) & \textit{a}, \textit{c}, \textit{o}, \textit{s} & \begin{tabular}[c]{@{}l@{}}\texttt{[('NULL', 'food quality', 'delicious', 'positive'),} \\ \texttt{\  ('waiter', 'service speed', 'way too slow', 'negative')]}\end{tabular} \\
\rowcolor{gray!5}
& \begin{tabular}[c]{@{}l@{}}Aspect-Category-Opinion-Sentiment \\ Quad Extraction (ACOS)\end{tabular} & \textit{a}, \textit{c}, \textit{o}, \textit{s} & \begin{tabular}[c]{@{}l@{}}\texttt{[('NULL', 'food quality', 'delicious', 'positive'),}\\ \texttt{\  ('waiter', 'service speed', 'way too slow', 'negative')]}\end{tabular} \\
\bottomrule
\end{tabular}
}
\vspace{0.3em}
\caption{\textbf{Overview of ABSA tasks supported by AnnoABSA.} For each task, a gold label is presented for the sentence \textit{``It was really delicious, but the waiter was way too slow''}. Aspect categories are commonly selected from a predefined set (\{food quality, service speed, ...\}). Notation: \textit{a} = aspect term, \textit{c} = aspect category, \textit{o} = opinion term, \textit{s} = sentiment polarity. In case of an implicit aspect, aspect term \textit{a} is set to \texttt{'NULL'}. The listed tasks are equivalent to those reported in the literature review by \citet{zhang2022survey}}
\label{tab:absa_tasks}
\end{table*}
\clearpage

\section{CLI Flags}\label{appendix:flags}

\begin{table}[ht]
\centering
\vspace{0.5em}
\resizebox{\textwidth}{!}{
\renewcommand{\arraystretch}{1.2}
\Large
\begin{tabular}{@{}l p{12.5cm} p{8cm}@{}}
\toprule
\textbf{Option} & \textbf{Description} & \textbf{Default} \\
\midrule
\multicolumn{3}{@{}l}{\textit{Server Configuration}} \\
\addlinespace[0.2em]
\rowcolor{gray!5}
\texttt{--backend} & Start only backend server & -- \\
\texttt{--backend-port} & Backend server port & \texttt{8000} \\
\rowcolor{gray!5}
\texttt{--frontend-port} & Frontend server port & \texttt{3000} \\
\texttt{--backend-ip} & Backend server IP address & \texttt{localhost} \\
\rowcolor{gray!5}
\texttt{--frontend-ip} & Frontend server IP address & \texttt{localhost} \\
\addlinespace[0.3em]
\hline
\multicolumn{3}{@{}l}{\textit{Session Management}} \\
\addlinespace[0.2em]
\texttt{--session-id} & Unique identifier for an annotation session. & \texttt{None} \\
\addlinespace[0.3em]
\hline
\multicolumn{3}{@{}l}{\textit{Annotation Elements}} \\
\addlinespace[0.2em]
\rowcolor{gray!5}
\texttt{--elements} & Sentiment elements to annotate & aspect\_term, aspect\_category, sentiment\_polarity, opinion\_term \\
\texttt{--polarities} & Valid sentiment polarities & positive, negative, neutral \\
\rowcolor{gray!5}
\texttt{--categories} & Valid aspect categories & Restaurant domain (13 categories) \\
\texttt{--implicit-aspect} & Allow implicit aspect terms & Enabled \\
\rowcolor{gray!5}
\texttt{--disable-implicit-aspect} & Disable implicit aspect terms & Disabled \\
\texttt{--implicit-opinion} & Allow implicit opinion terms & Disabled \\
\rowcolor{gray!5}
\texttt{--disable\_implicit\_opinion} & Disable implicit opinion terms & Enabled \\
\addlinespace[0.3em]
\hline
\multicolumn{3}{@{}l}{\textit{Interface and Processing}} \\
\addlinespace[0.2em]
\texttt{--disable\_clean\_phrases} & Disable automatic punctuation cleaning from phrase start/end & Enabled \\
\rowcolor{gray!5}
\texttt{--disable-save-positions} & Disable saving phrase positions & Enabled \\
\texttt{--disable-click-on-token} & Disable click-on-token feature & Enabled \\
\rowcolor{gray!5}
\texttt{--auto-positions} & Enable automatic position filling on startup & Disabled \\
\texttt{--annotation-guidelines} & Path to PDF file containing annotation guidelines & Disabled \\
\addlinespace[0.3em]
\hline
\multicolumn{3}{@{}l}{\textit{Analytics and Timing}} \\
\addlinespace[0.2em]
\texttt{--store-time} & Store timing data for annotation sessions & Disabled \\
\rowcolor{gray!5}
\texttt{--display-avg-annotation-time} & Display average annotation time in the interface & Disabled \\
\addlinespace[0.3em]
\hline
\multicolumn{3}{@{}l}{\textit{AI Integration}} \\
\addlinespace[0.2em]
\rowcolor{gray!5}
\texttt{--ai-suggestions} & Enable AI-powered prediction suggestions using LLM & Disabled \\
\texttt{--disable-ai-automatic-prediction} & Disable automatic AI prediction triggering & Disabled \\
\rowcolor{gray!5}
\texttt{--llm-model} & LLM employed for suggestions (e.g., \texttt{gemma3:4b}, \texttt{gpt-4o}) & \texttt{gemma3:4b} \\
\texttt{--openai-key} & OpenAI API key for using OpenAI models & -- \\
\rowcolor{gray!5}
\texttt{--n-few-shot} & Maximum number of few-shot examples in LLM context & \texttt{10} \\
\addlinespace[0.3em]
\hline
\multicolumn{3}{@{}l}{\textit{Configuration Management}} \\
\addlinespace[0.2em]
\texttt{--save-config} & Save config to JSON file & -- \\
\rowcolor{gray!5}
\texttt{--show-config} & Display current configuration in console & -- \\
\bottomrule
\end{tabular}
}
\vspace{0.3em}
\caption{\textbf{CLI options for AnnoABSA.} AnnoABSA offers extensive customization across server settings, annotation logic, and AI integration.}
\label{tab:cli-options}
\end{table}

\clearpage

\section{Prompt for RAG-Based Suggestions}\label{appendix:prompt}

\begin{figure}[ht]
    \centering
    \includegraphics[width=\textwidth]{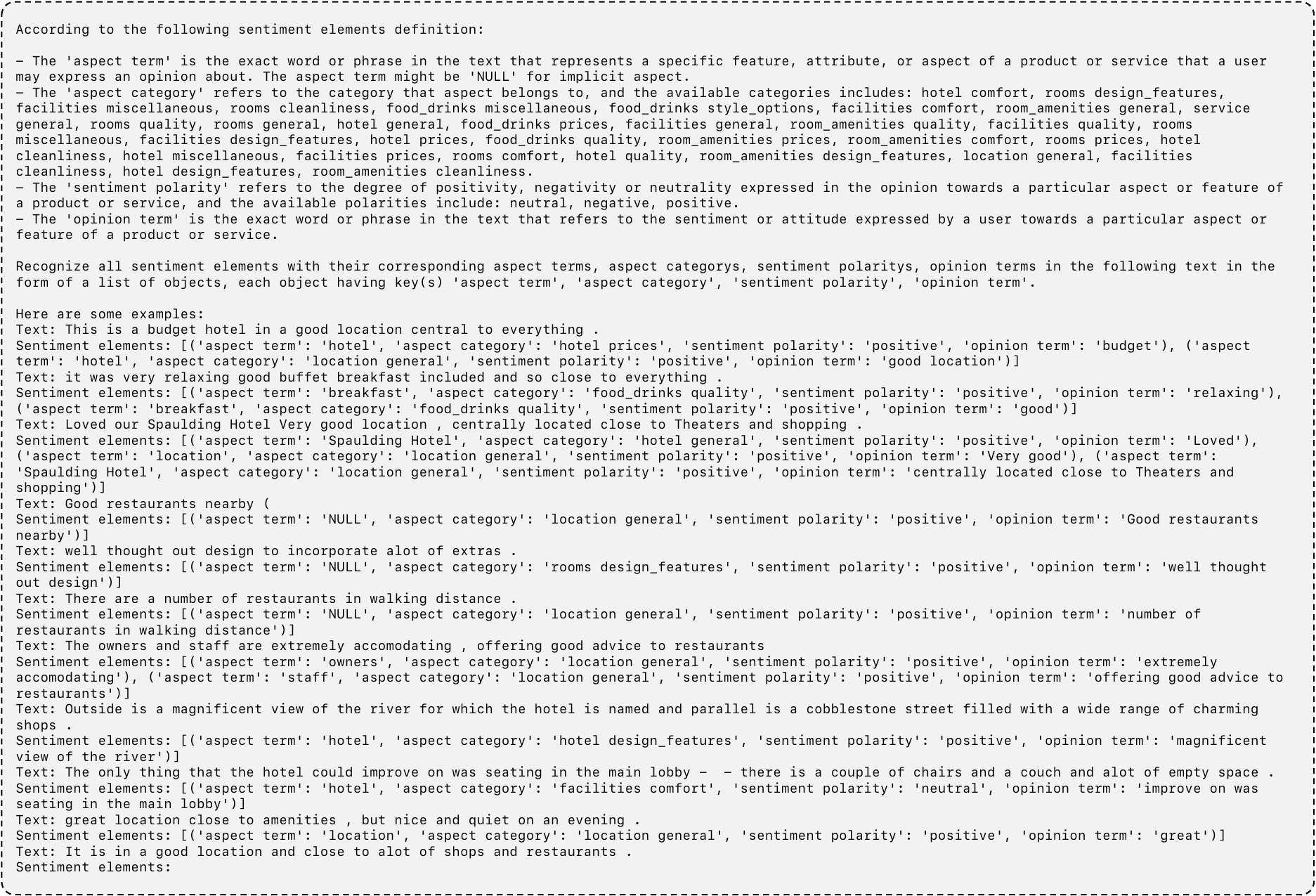}
    \caption{\textbf{Prompt used for RAG-based suggestion prediction.} 
The prompt includes a task description with explanations of sentiment elements, 
ten in-context demonstrations, and the target text for aspect prediction. The few-shot examples shown are taken from the Coursera dataset and include annotations for ASQP}
\label{fig:prompt-example}
\end{figure}

\end{document}